\documentclass[10pt,twocolumn,letterpaper]{article}

\usepackage[pagenumbers]{wacv}

\usepackage{graphicx}
\usepackage{amsmath}
\usepackage{amssymb}
\usepackage{booktabs}
\usepackage{tabularx}
\usepackage[table]{xcolor}
\usepackage{fontawesome}
\usepackage{multirow}
\usepackage{array}

\newcolumntype{P}[1]{>{\centering\arraybackslash}p{#1}}
\newcommand{\figref}[1]{Figure~\ref{fig:#1}}
\newcommand{\tabref}[1]{Table~\ref{tabularx:#1}}

\usepackage[pagebackref,breaklinks,colorlinks]{hyperref}

\usepackage[capitalize]{cleveref}
\crefname{section}{Sec.}{Secs.}
\Crefname{section}{Section}{Sections}
\Crefname{table}{Table}{Tables}
\crefname{table}{Tab.}{Tabs.}

\def\confName{arXiv}
\def\confYear{2026}

\begin{document}

%%%%%%%%% TITLE
\title{A Universal Action Space for General Behavior Analysis}

\author{
Hung-Shuo Chang$^{1}$,
Yue-Cheng Yang$^{1}$,
Yu-Hsi Chen$^{1}$, \\
Wei-Hsin Chen$^{2}$,
Chien-Yao Wang$^{1}$,
James C. Liao$^{3}$, \\
Chien-Chang Chen$^{2}$,
Hen-Hsen Huang$^{1}$,
Hong-Yuan Mark Liao$^{1}$ \\
$^{1}$Institute of Information Science, Academia Sinica, Taiwan \\
{\tt\small \{jonathanc, skyworker, franktpmvu, kinyiu, hhhuang, liao\}@iis.sinica.edu.tw} \\
$^{2}$Institute of Biomedical Sciences, Academia Sinica, Taiwan \\
{\tt\small \{vic7538, ccchen\}@ibms.sinica.edu.tw} \\
$^{3}$Institute of Biological Chemistry, Academia Sinica, Taiwan \\
{\tt\small liaoj@gate.sinica.edu.tw} \\
}

\maketitle

% \vspace*{-18px}

% Input arXiv-ready sections (all in English)
\begin{abstract}
    Analyzing animal and human behavior has long been a challenging task in computer vision. Early approaches from the 1970s to the 1990s relied on hand-crafted edge detection, segmentation, and low-level features such as color, shape, and texture to locate objects and infer their identities—an inherently ill-posed problem. Behavior analysis in this era typically proceeded by tracking identified objects over time and modeling their trajectories using sparse feature points, which further limited robustness and generalization. A major shift occurred with the introduction of ImageNet by Deng and Li in 2010, which enabled large-scale visual recognition through deep neural networks and effectively served as a comprehensive visual dictionary. This development allowed object recognition to move beyond complex low-level processing toward learned high-level representations. In this work, we follow this paradigm to build a large-scale Universal Action Space (UAS) using existing labeled human-action datasets. We then use this UAS as the foundation for analyzing and categorizing mammalian and chimpanzee behavior datasets.
    The source code is released on GitHub at \url{https://github.com/franktpmvu/Universal-Action-Space}.
\end{abstract}

\section{Introduction}

    \begin{figure*}[t]
        \centering
        \includegraphics[width = .9\linewidth]{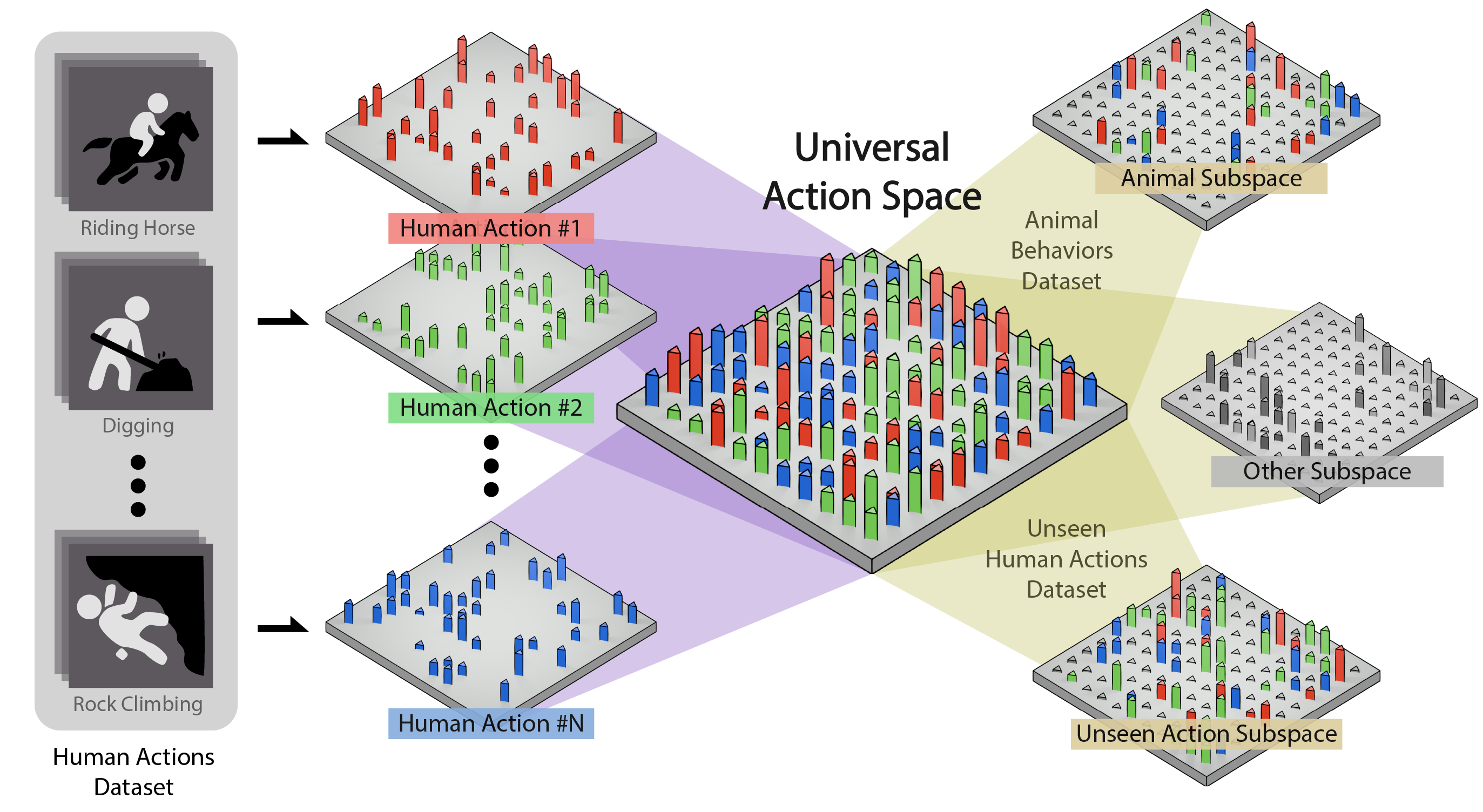}
        \vspace{-8pt}
        \caption{
            An overview of the Universal Action Space (UAS). We first derive features from complex human actions to build the UAS, and downstream tasks then utilize this feature space to construct their respective subspaces.
        }
        \label{fig:uas_overview}
        \vspace{-12pt}
    \end{figure*}
    
    Action recognition has been a core research topic in computer vision, aiming to extract spatio-temporal features from video data to support high-level applications such as human–computer interaction, sports analysis, and animal behavior studies. Since actions inherently involve both spatial posture patterns and temporal dynamics, accurately recognizing actions in complex environments remains a challenging research problem.

    Early methods relied on hand-crafted features and classification rules, using manually designed descriptors to identify behavior patterns in videos~\cite{laptev2003,wang2011action,wang2013action}. These approaches not only required extensive domain expertise but were also highly sensitive to background noise, viewpoint variations, and temporal scale differences, severely limiting their generalization and applicability to downstream tasks.

    With the rise of deep learning, progress in action recognition has been driven by large-scale supervised datasets~\cite{deng2009imagenet, kuehne2011hmdb, soomro2012ucf101dataset101human, kay2017kineticshumanactionvideo} and deep action recognition models~\cite{simonyan2014,wang2015action,du20153d,joao2017i3d,gammulle2017twostream,ullah2018action,ma2019ts}. 
    \iffalse
    On the data side, ImageNet~\cite{deng2009imagenet} enabled transferable visual representations through large-scale image pretraining, while video benchmarks such as HMDB51~\cite{kuehne2011hmdb}, UCF101~\cite{soomro2012ucf101dataset101human}, and Kinetics~\cite{kay2017kineticshumanactionvideo} provided labeled data for learning temporal dynamics in videos. In parallel, deep action recognition models~\cite{simonyan2014,wang2015action,du20153d,joao2017i3d,gammulle2017twostream,ullah2018action,ma2019ts} were developed to learn discriminative spatio-temporal features in a data-driven manner, substantially improving recognition performance.
    \fi
    Despite these advances, modern action recognition models remain heavily dependent on large-scale pretraining: models trained on foundation datasets and then adapted to downstream tasks often achieve excellent results under various fine-tuning strategies~\cite{Chen2022MMViT,arnab2021vivit,gberta_2021_ICML,yang2022recurring}. However, such pretraining and adaptation typically require substantial compute and memory, which can be prohibitive for research teams with limited resources.

    In~\cite{chen2025deeplearningbasedanimalbehavior}, we proposed the Universal Action Space (UAS) trained on Kinetics-600, providing a transferable representation capable of modeling diverse animal behaviors without backbone fine-tuning. Given the compositional complexity of human behaviors, we hypothesize that a UAS trained on such a dataset inherently captures latent subspaces corresponding to a wide spectrum of animal behaviors, even those not explicitly present in the training data.

    Building on this insight, we adopt UAS as a frozen backbone and project animal behaviors into this pretrained representation space, thereby forming behavior-specific subspaces. In this framework, UAS serves as a versatile foundation representation supporting multiple downstream tasks, where only a lightweight classifier is trained atop the frozen UAS, eliminating the need for backbone pretraining or fine-tuning. This approach substantially reduces the computational resources and training time typically required for large-scale backbone models, making it particularly advantageous for research teams with limited resources. The overall UAS framework is illustrated in~\figref{uas_overview}.

    The main contributions of this paper are:
    \begin{itemize}
    \item We demonstrate that a Universal Action Space (UAS) pretrained on complex human actions transfers effectively to diverse downstream action recognition tasks.
    \item We demonstrate that UAS enables parameter- and time-efficient adaptation by training only lightweight heads on a frozen backbone, substantially reducing training cost under limited compute.
    \end{itemize}
\section{Related Works}
    
    \subsection{Action Recognition and Behavior Analysis}
    
        Early approaches to action recognition primarily relied on traditional image processing and computer vision techniques, employing hand-crafted features to represent action content. These methods typically detected feature points on moving human or animal bodies and tracked them over time, followed by recognition using manually designed classification rules or statistical models~\cite{laptev2003,wang2011action,wang2013action,pai2004pedestrian,liang2009learning,chen2015basketball,tang2015robust}.

        Representative examples include the work of Pai et al~\cite{pai2004pedestrian}, who developed a pedestrian detection system for intersections that combined foreground extraction with dynamic graph matching to improve multi-person tracking accuracy. Similarly, Liang et al.~\cite{liang2009learning} proposed modeling fine-grained human actions using Variable-length Markov Models (VLMMs), borrowing ideas from language modeling to systematically describe human motion. However, the limited modeling capacity of early statistical approaches and the computational resources available at the time constrained their effectiveness.

        In~\cite{chen2015basketball}, Chen et al. analyzed basketball strategies by projecting players' trajectories onto a tactical board, deriving motion templates from velocity features to enable real-time recognition of offensive plays. Tang et al.~\cite{tang2015robust} leveraged the Microsoft Kinect to capture multimodal information—RGB sequences, skeletal data, and point clouds—within a short range, and further proposed a Kinect-based dataset to bridge 2D RGB inputs with corresponding 3D skeletal and point cloud data, extending applicability to longer-range activities.

        At the same time, researchers sought more generic motion descriptors. Laptev et al.~\cite{laptev2003} extended Harris and Förstner interest point operators into the spatio-temporal domain, introducing scale-invariant descriptors for event classification. Wang et al.~\cite{wang2011action,wang2013action} advanced this direction with dense trajectory methods, using optical flow to track points across frames and extracting shape, motion, and boundary features for robust action representation.

        While influential, these traditional approaches shared several fundamental limitations. First, the equipment required to capture rich 3D motion data was often expensive and difficult to deploy. Moreover, the lack of GPU-class computational power limited real-time analysis. Lastly, the use of sparse feature points sacrificed fine-grained muscular motion, resulting in poor accuracy for detailed behavior analysis. These challenges ultimately underscored the need for data-driven methods that could automatically learn spatio-temporal representations from raw data.

    \subsection{Deep Learning for Action Recognition}
    
        With the rise of deep learning, Convolutional Neural Networks (CNNs) became the dominant paradigm in computer vision. To adapt CNNs for action recognition, researchers proposed architectures that explicitly model temporal dynamics and motion information in videos~\cite{simonyan2014,du20153d,joao2017i3d,ma2019ts}. Since actions evolve over time, effective recognition requires joint reasoning over both spatial appearance and temporal evolution.

        A representative approach is the two-stream framework, which employs separate CNNs to process RGB frames and optical flow, followed by fusion of the two prediction streams~\cite{simonyan2014,wang2015action}. Although effective, this late-fusion strategy limits fine-grained interactions between appearance and motion features, while dependence on precomputed optical flow can make performance sensitive to noise. Moreover, two-stream architectures struggle to capture long-range temporal dependencies.

        To model spatio-temporal patterns more directly, CNNs were extended from 2D to 3D~\cite{du20153d,joao2017i3d}, enabling convolutional filters to operate jointly across space and time. While 3D CNNs are effective at capturing short-term motion, they significantly increase computational cost and parameter count, and their temporal receptive field remains inherently limited. Another line of work combines CNN-based feature extractors with recurrent architectures such as Long Short-Term Memory (LSTM) networks~\cite{gammulle2017twostream,ullah2018action,ma2019ts} to model longer-term temporal structure. However, the sequential nature of recurrent processing reduces parallelism, making such models less efficient and more difficult to scale.

        More recently, Transformer-based models have been introduced for video understanding, inspired by the success of Vision Transformers (ViT) in image recognition~\cite{dosovitskiy2020vit} and their extensions to action recognition~\cite{Chen2022MMViT,arnab2021vivit,gberta_2021_ICML,yang2022recurring}. Unlike CNN-based approaches that incorporate temporal reasoning via auxiliary branches or 3D convolutions, Transformers model spatio-temporal tokens directly using self-attention, enabling richer interactions between appearance and motion cues. Furthermore, self-attention supports parallel computation during training, offering improved scalability compared to recurrent formulations.

    \subsection{Foundation Datasets and Universal Representations}
    
        The introduction of ImageNet~\cite{deng2009imagenet} ushered in the era of deep learning. Through supervised learning on large-scale annotated data, models can effectively learn highly discriminative visual features, forming a "visual dictionary." Under this paradigm, recognizing objects in images becomes a "dictionary lookup" task. Converting the image recognition problem into visual dictionary lookup helps avoid many ill-posed vision problems (such as edge detection or image segmentation). Because ImageNet can be widely applied to image recognition problems, it has naturally become one of the most important foundation models in computer vision.

        Inspired by ImageNet's success, subsequent research extended this concept to action recognition tasks. Researchers collected and annotated large volumes of videos to establish datasets with diverse action categories, forming "action dictionaries." Representative human action datasets include HMDB51~\cite{kuehne2011hmdb}, UCF101~\cite{soomro2012ucf101dataset101human}, and Kinetics~\cite{kay2017kineticshumanactionvideo}. These datasets contain abundant human daily behaviors and athletic actions, providing models with rich and diverse action features.

        Beyond human-centric action datasets, recent studies have increasingly focused on animal behavior analysis, leading to the introduction of several large-scale animal action datasets~\cite{Ng_2022_CVPR,Chen_2023_CVPR}. From the perspective of action complexity and behavioral diversity, human actions exhibit the richest and most diverse spatio-temporal patterns among animal species. This observation motivates a fundamental question: can an action space with maximal complexity and high dimensionality be learned from human actions and behaviors?
        We hypothesize that such a representation, when trained as a foundation model, can serve as a universal action space. Within this space, actions and behaviors of other animal species would correspond to lower-dimensional subspaces, activating only a subset of the learned dimensions. By leveraging this shared representation, a unified framework for recognizing both human and animal actions and behaviors becomes possible.

    \subsection{Universal Action Space}
    
        The concept of a Universal Action Space (UAS) was introduced by our team in 2025~\cite{chen2025deeplearningbasedanimalbehavior}. In that work, we pretrained a Microsoft Video Swin Transformer (VST)~\cite{liu2021video} on the Kinetics-600 dataset~\cite{kay2017kineticshumanactionvideo,carreira2018shortnotekinetics600} and defined the resulting high-dimensional action embedding as the UAS. This pretrained representation was subsequently adopted as a foundation model for downstream mouse behavior analysis, with a particular focus on identifying pain-related behavioral patterns.
        Experimental results demonstrated that action representations learned from complex human behaviors can induce simpler, lower-dimensional behavioral subspaces for animals, enabling effective transfer to animal behavior analysis tasks.
    
\section{Proposed Methods}

    \subsection{Definition of UAS}
    
        The Kinetics dataset~\cite{kay2017kineticshumanactionvideo}, introduced by DeepMind in 2017, is a large-scale human action video dataset initially comprising 400 action classes, later expanded to 600 in Kinetics-600~\cite{carreira2018shortnotekinetics600}. Each video clip, approximately 10 seconds long, is sourced from diverse YouTube videos and represents a wide variety of human activities (e.g., riding horse, petting cat). Owing to its extensive coverage of motion patterns and rich temporal diversity, we hypothesize that an action space trained on the Kinetics dataset can inherently capture a wide range of behavioral representations, extending beyond human actions to encompass various forms of animal behavior as well.

        The UAS proposed in~\cite{chen2025deeplearningbasedanimalbehavior} provides a transferable representation capable of modeling diverse animal behaviors without requiring task-specific backbone pretraining. Given the compositional richness of human behaviors, a UAS learned from large-scale human action datasets implicitly encodes latent subspaces corresponding to a broad range of animal behaviors. In our framework, the UAS is employed as a frozen backbone, and animal behaviors are projected into this pretrained representation space, forming behavior-specific subspaces.

        By eliminating the need for large-scale backbone retraining, the proposed UAS substantially reduces computational cost and training time, making it particularly well suited for research settings with limited computational resources.
    
    \subsection{Construction of UAS}
    
        Extracting motion features from human action videos constitutes the first step in constructing the UAS. To achieve this, we employ the Video Swin Transformer (VST)~\cite{liu2021video}, a state-of-the-art model introduced by Microsoft in 2021 for video-based action extraction. VST captures motion features through shifted windows that compute spatial activation maps (heat maps) for each frame, which are then sequentially processed along the temporal axis. The temporal evolution of these heat maps effectively encodes the dynamics of motion across frames.

\begin{figure}[htbp]
        \centering
        \includegraphics[width = 1.\linewidth]{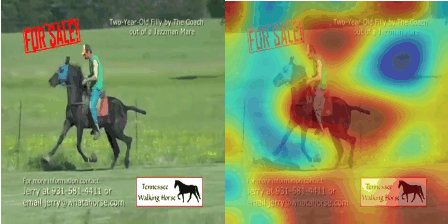}
        \vspace{-8pt}
        \caption{
            Heat map visualization of motion features captured by the Video Swin Transformer. Regions with higher motion intensity are shown in red.
        }
        \label{fig:vst_heatmap}
    \end{figure}

        As shown in~\figref{vst_heatmap}, redder regions in the extracted heat maps correspond to areas of higher motion intensity within a frame. Following the procedure described in~\cite{chen2025deeplearningbasedanimalbehavior}, we use VST to extract motion features from all 600 human action categories in Kinetics-600 and project them into the UAS, completing its training process as illustrated in the central portion of~\figref{uas_overview}.
    
    \subsection{Training Domain-Specific Subspace via UAS}
    
    Once constructed, the UAS serves as a universal foundational representation, providing transferable action features across diverse downstream tasks. For each downstream task, we project its representations to learn the corresponding domain-specific subspaces. As illustrated on the right side of~\figref{uas_overview}, the UAS transfers feature representations into these task-specific subspaces, with each subspace capturing behavior patterns and features relevant to its particular application domain. In this framework, the UAS functions as a shared foundation from which multiple downstream domains can be effectively represented and analyzed.

    In this paper, we evaluate the effectiveness of the UAS using animal action classification tasks as downstream applications. For each input video, the Video Swin Transformer (VST) first extracts animal behavior features and maps them into the UAS, which is constructed from large-scale human action data. Through this mapping, animal actions are embedded into the high-dimensional UAS and represented as task-specific subspaces. We then employ a linear classifier~\cite{alain2018understandingintermediatelayersusing} on top of the frozen UAS features to perform downstream animal action classification.
    
\section{Experiments}
    
    \begin{table*}[h]
        \begin{center}
            \begin{tabular}{ll}
            \scalebox{.98}
            {
                \setlength\tabcolsep{0pt}
                \begin{tabularx}{1.00\linewidth}{P{0.16\linewidth}|P{0.13\linewidth}P{0.09\linewidth}P{0.16\linewidth}|P{0.08\linewidth}P{0.08\linewidth}|P{0.18\linewidth}P{0.12\linewidth}}  
                \toprule
                
                model & Backbone & Pre-trained & Training Strategy & Top-1 \color{blue}{$\uparrow$} & MCA \color{blue}{$\uparrow$} & Training Time (hr) \color{red}{$\downarrow$} & \#Params (K)\color{red}{$\downarrow$} \\
                
                \midrule
                
                MammalNet~\cite{chen2023mammalnet} & MViTv2~\cite{li2022mvitv2} & K-400 & Full Fine-tune & 46.6 & 37.8 & 248.8 & 51,028.7 \\
                
                \midrule
                
                Ours & VST & K-400 & Linear Probe & \textbf{56.6} & \textbf{43.2} & \textbf{8.3} & \textbf{12.3} \\
                
                \bottomrule
                \end{tabularx}
            }
            \end{tabular}
        \end{center}
        \vspace{-6pt}
        \caption{
            Experimental results on the MammalNet dataset. K-400 denotes Kinetics-400. Our UAS-based approach achieves 21.5\% higher top-1 accuracy while using 30× less training time and 4150× fewer parameters.
        }
        \label{tabularx:mammalnet_result}
    \end{table*}
    
    \begin{table*}[h]
        \begin{center}
            \begin{tabular}{ll}
            \scalebox{.98}
            {
                \setlength\tabcolsep{0pt}
                \begin{tabularx}{1.00\linewidth}{P{0.16\linewidth}|P{0.13\linewidth}P{0.09\linewidth}P{0.16\linewidth}|P{0.08\linewidth}P{0.08\linewidth}|P{0.18\linewidth}P{0.12\linewidth}}
                \toprule
                
                model & Backbone & Pre-trained & Training Strategy & Top-1 \color{blue}{$\uparrow$} & MCA \color{blue}{$\uparrow$} & Training Time (hr) \color{red}{$\downarrow$} & \#Params (K)\color{red}{$\downarrow$} \\
                
                \midrule
                
                ChimpBehave~\cite{fuchs2024chimpbehave} & X3D~\cite{feichtenhofer2020x3d} & K-400 & Full Fine-tune & 90.3 & 67.2 & - & 6,153.4 \\
                
                \midrule
                
                Ours & VST & K-400 & Linear Probe & 93.7 & 65.8 & \textbf{3.9} & \textbf{7.2} \\
                Ours & VST & K-600 & Linear Probe & 93.5 & \textbf{72.3} & \textbf{3.9} & \textbf{7.2} \\
                Ours & VST & K-700 & Linear Probe & \textbf{94.2} & 56.4 & \textbf{3.9} & \textbf{7.2} \\
                
                \bottomrule
                \end{tabularx}
            }
            \end{tabular}
        \end{center}
        \vspace{-6pt}
        \caption{
            Experimental results on the ChimpBehave dataset. K-400, K-600, and K-700 denote Kinetics-400, Kinetics-600, and Kinetics-700, respectively. UAS achieves up to 3.8\% higher top-1 accuracy and 7.6\% higher MCA while using 854× fewer parameters.
        }
        \label{tabularx:chimpbehave_result}
    \end{table*}
    
    \begin{figure*}[t]
        \centering
        \includegraphics[width = .9\linewidth]{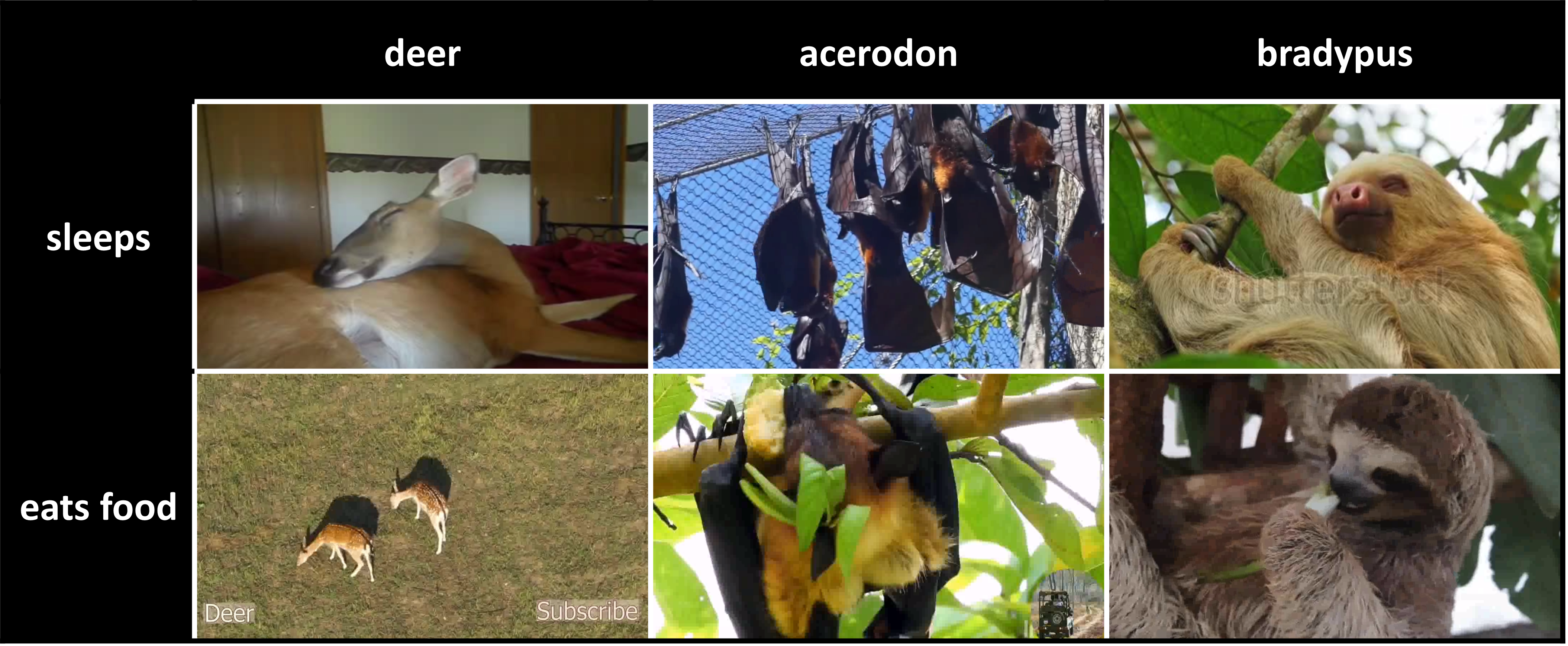}
        \vspace{-8pt}
        \caption{
            Examples of mammalian behaviors included in the MammalNet dataset. This figure demonstrates the diversity of actions across different species.
        }
        \label{fig:mammal_diff_action}
    \end{figure*}
    
    \begin{table*}[h]
        \begin{center}
            \begin{tabular}{ll}
            \scalebox{.98}
            {
                \setlength\tabcolsep{0pt}
                \begin{tabularx}{0.77\linewidth}{P{0.16\linewidth}|P{0.10\linewidth}P{0.10\linewidth}P{0.10\linewidth}|P{0.18\linewidth}P{0.13\linewidth}}
                \toprule
                
                Training Strategy & Top-1 \color{blue}{$\uparrow$} & Top-5 \color{blue}{$\uparrow$} & MCA \color{blue}{$\uparrow$} & Training Time (hr) \color{red}{$\downarrow$} & \#Params (K)\color{red}{$\downarrow$} \\
                
                \midrule
                
                Full Fine-tune & \textbf{88.8} & 98.0 & \textbf{88.8} & 105 & 87,744.6 \\
                LoRA & 88.6 & \textbf{98.1} & 88.6 & 86 & 1,645.7 \\
                Linear Probe (Ours) & 87.9 & 97.8 & 87.8 & \textbf{54} & \textbf{105.6} \\
                
                \bottomrule
                \end{tabularx}
            }
            \end{tabular}
        \end{center}
        \vspace{-6pt}
        \caption{
            Experimental results on the Kinetics-700 diff set (103 action classes unseen in Kinetics-600). Our UAS-based linear probing achieves competitive accuracy (\text{-}0.9\% vs. full fine-tuning) while using 51\% training time and 0.12\% parameters.
        }
        \label{tabularx:k700diff_performance}
    \end{table*}
    
    \subsection{Experimental Settings}

        For our experiments, we use an Inflated 3D ConvNet (I3D)~\cite{carreira2017i3d} head as the linear classifier, consisting of a single average pooling layer followed by a fully connected layer. With the UAS frozen, this lightweight design minimizes trainable parameters and training time. High accuracy across different domains with such compact classifiers demonstrates the efficiency and robustness of the UAS foundation. We evaluate on the following datasets:    
        \begin{itemize}
            \item \textbf{MammalNet}~\cite{chen2023mammalnet}: 173 mammalian species with 12 behavior categories; 38,296K annotated frames in total.
            \item \textbf{ChimpBehave}~\cite{fuchs2024chimpbehave}: 7 chimpanzee behavior classes; 193K annotated frames in total.
        \end{itemize}

        We use two common evaluation metrics in action recognition tasks to assess model performance:
        \begin{itemize}
            \item \textbf{Top-1 Accuracy}: Measures the proportion of correctly classified instances across the entire test set, assigning equal weight to each instance. This metric directly reflects the overall predictive performance of the model but can be sensitive to class imbalance.
            \item \textbf{Mean Class Accuracy}: Computes classification accuracy independently for each class and then averages across all classes, assigning equal weight to each class. This metric provides a more balanced evaluation in the presence of imbalanced class distributions.
        \end{itemize}

        Additionally, to further demonstrate UAS's efficiency, we report the \textbf{total training time} and \textbf{total number of trainable parameters} during the training phase to analyze training performance.

    \subsection{Results on MammalNet}
    
        The results on the MammalNet dataset are reported in~\tabref{mammalnet_result}. We compare our method against the MammalNet baseline, which adopts MViTv2~\cite{li2022mvitv2} as the backbone encoder. Our UAS-based variant employs the Microsoft Video Swin Transformer (VST)~\cite{liu2021video} as the encoder. Both models are pretrained on Kinetics-400. The baseline is trained via full fine-tuning, whereas our approach uses linear probing on top of a frozen UAS backbone.

        Compared with the baseline, the proposed UAS framework achieves a substantial performance improvement, yielding a \textbf{21.5\%} increase in Top-1 accuracy and a \textbf{14.3\%} increase in Mean Class Accuracy. In terms of computational efficiency, UAS requires only \textbf{3.3\%} of the baseline’s total training time, and the number of trainable parameters is reduced to just \textbf{0.02\%} of the baseline. These results demonstrate that UAS not only improves classification performance but also dramatically reduces computational and training costs.

        We investigate the reasons behind the relatively low Top-1 accuracy (56.6\%) observed on the MammalNet dataset. Our analysis indicates that the primary limiting factor is insufficient data volume relative to task complexity. MammalNet comprises 173 mammalian species and 12 distinct behavior categories, resulting in a total of $173 \times 12 = 2{,}076$ fine-grained classes. Under such high class cardinality, limited training samples make it difficult for decision hyperplanes in the feature space to effectively separate different species-action combinations. Consequently, the achievable Top-1 accuracy is constrained.

    \subsection{Results on ChimpBehave}
    
        \tabref{chimpbehave_result} presents the results on the ChimpBehave dataset. In this experiment, we compare three Kinetics-series datasets as foundations. The baseline method uses X3D~\cite{feichtenhofer2020x3d} as its backbone, pretrained on Kinetics-400, with full fine-tuning as the training strategy. Our method employs the Video Swin Transformer (VST) as the backbone and adopts linear probing for training, with the backbone pretrained on Kinetics-400, Kinetics-600, and Kinetics-700, respectively.

        The results show that all UAS-based configurations consistently outperform the baseline in Top-1 accuracy, with improvements of at least \textbf{3.8\%}. Among them, the UAS pretrained on Kinetics-600 achieves the highest Mean Class Accuracy, yielding a \textbf{7.6\%} improvement over the baseline. In terms of computational efficiency, all three UAS variants require only \textbf{0.12\%} of the baseline’s trainable parameters, further demonstrating the efficiency and effectiveness of the proposed UAS framework.

        Notably, the Top-1 accuracy on ChimpBehave exceeds 93\% across methods. This high performance can be attributed to the relatively low task complexity of ChimpBehave, which contains videos of a single animal species (chimpanzees) and only seven behavior categories. Under this setting, each behavior class is supported by sufficient training data, enabling the backbone networks to learn highly discriminative representations.

    \subsection{Discussion}
    
        The above experimental results demonstrate that UAS not only improves classification performance on animal action recognition tasks but also substantially reduces model training time and the number of trainable parameters, highlighting its practical utility as a foundation representation.

        On the MammalNet dataset, disregarding class imbalance, each action category contains on average approximately 3,193K frames. However, behavioral patterns vary across species within the same action category, requiring action features to be further subdivided by species. As shown in~\figref{mammal_diff_action}, these inter-species differences are evident in the training set. Consequently, each species has on average only about 18K frames per action type for training. This high degree of data dispersion reduces the number of effective samples per category and compresses the representation of each action in the latent space.

        In contrast, the ChimpBehave dataset contains roughly 27K frames per action on average. Since all videos depict a single species, there is no need to account for cross-species variation, resulting in more training data per action category and more robust feature learning.

        Taken together, the results in~\tabref{mammalnet_result} and~\tabref{chimpbehave_result} indicate that ChimpBehave yields overall higher performance compared to MammalNet, consistent with the differences in data availability and species diversity.

    \subsection{Ablation Study}

        In the ablation study, we use Kinetics-600 as the foundation and select the difference set of Kinetics-700 relative to Kinetics-600 as the downstream action recognition task, which comprises 103 newly added human action categories~\cite{carreira2022shortnotekinetics700human}. Since all Kinetics-series datasets focus on human actions, similar to the ChimpBehave dataset, there are no cross-species considerations. Moreover, each action category contains at least 600 training videos of 5 seconds or longer (at least 72K frames), providing abundant data and a relatively balanced distribution. This setup minimizes the impact of data imbalance and allows the study to focus on comparing the effectiveness of different training strategies.

        The results, reported in~\tabref{k700diff_performance}, show that UAS-based linear probing achieves performance comparable to full fine-tuning and LoRA, indicating that the action subspace constructed by UAS already encodes highly discriminative representations. In terms of efficiency, linear probing requires only \textbf{51.4\%} and \textbf{62.8\%} of the training time of full fine-tuning and LoRA, respectively, while the number of trainable parameters is reduced to merely \textbf{0.12\%} and \textbf{6.4\%} of theirs. These results demonstrate that UAS can deliver comparable performance with substantially lower training cost.
        
\section{Conclusion}

    We explored a new deep learning–based framework for analyzing animal behavior from video, addressing a longstanding challenge in computer vision. By leveraging the human action dataset Kinetics-600 to construct a largescale Universal Action Space (UAS), we avoid training
    heavy backbones directly on animal data. Using this UAS as a shared representation, we successfully analyzed two downstream datasets—one of chimpanzee behaviors and one of diverse mammals—achieving strong recognition performance without collecting large additional video corpora.

%%%%%%%%% ACKNOWLEDGMENTS (Optional for arXiv)
% \section*{Acknowledgments}
% This work was supported by...

% \clearpage

%%%%%%%%% REFERENCES
{\small
\bibliographystyle{ieee_fullname}
\bibliography{egbib}
}

\end{document}